\documentclass[pdflatex, sn-nature]{sn-jnl}

\usepackage{graphicx}%
\usepackage{multirow}%
\usepackage{amsmath,amssymb,amsfonts}%
\usepackage{amsthm}%
\usepackage{mathrsfs}%
\usepackage[title]{appendix}%
\usepackage{xcolor}%
\usepackage{textcomp}%
\usepackage{manyfoot}%
\usepackage{booktabs}%
\usepackage{algorithm}%
\usepackage{algorithmicx}%
\usepackage{algpseudocode}%
\usepackage{listings}%
\usepackage{orcidlink}%
\usepackage[switch]{lineno}
\usepackage{fancyhdr}

\theoremstyle{thmstyleone}%
\theoremstyle{thmstyletwo}%

\theoremstyle{thmstylethree}%

\begin{document}


\title[Article Title]{Leveraging large language models for nano synthesis mechanism explanation: solid foundations or mere conjectures?}


\author[1,2]{\fnm{Yingming} \sur{Pu}}

\author[3]{\fnm{Liping} \sur{Huang}}

\author*[2,5]{\fnm{Tao} \sur{Lin} \orcidlink{0000-0002-3246-6935}}\email{lintao@westlake.edu.cn}

\author*[3,4]{\fnm{Hongyu} \sur{Chen} \orcidlink{0000-0002-5325-9249}}\email{chenhongyu@westlake.edu.cn}

 \affil[1]{\orgname{Zhejiang University}, \orgaddress{\city{Hangzhou}, \postcode{310030}, \state{Zhejiang}, \country{China}}}

\affil[2]{\orgdiv{Department of Computer Science and Engineering, School of Engineering}, \orgname{Wetlake University}, \orgaddress{\city{Hangzhou}, \postcode{310030}, \state{Zhejiang}, \country{China}}}

\affil[3]{\orgdiv{Department of Chemistry, School of Science and Key Laboratory for Quantum Materials of Zhejiang Province}, \orgname{Wetlake University}, \orgaddress{ \city{Hangzhou}, \postcode{310030}, \state{Zhejiang}, \country{China}}}

\affil[4]{\orgdiv{Institute of Natural Sciences}, \orgname{Westlake Institute for Advanced Study}, \orgaddress{\city{Hangzhou}, \postcode{310024}, \country{China}}}

\affil[5]{\orgdiv{Research Center for Industries of the Future}, \orgname{Wetlake University}, \orgaddress{\city{Hangzhou}, \postcode{310030}, \state{Zhejiang}, \country{China}}}


%
%
%


\abstract{With the rapid development of artificial intelligence (AI), large language models (LLMs) such as GPT-4 have garnered significant attention in the scientific community, demonstrating great potential in advancing scientific discovery. This progress raises a critical question: are these LLMs well-aligned with real-world physicochemical principles? Current evaluation strategies largely emphasize fact-based knowledge, such as material property prediction or name recognition, but they often lack an understanding of fundamental physicochemical mechanisms that require logical reasoning. To bridge this gap, our study developed a benchmark consisting of 775 multiple-choice questions focusing on the mechanisms of gold nanoparticle synthesis. By reflecting on existing evaluation metrics, we question whether a direct true-or-false assessment merely suggests conjecture. Hence, we propose a novel evaluation metric, the confidence-based score (c-score), which probes the output logits to derive the precise probability for the correct answer. Based on extensive experiments, our results show that in the context of gold nanoparticle synthesis, LLMs understand the underlying physicochemical mechanisms rather than relying on conjecture. This study underscores the potential of LLMs to grasp intrinsic scientific mechanisms and sets the stage for developing more reliable and effective AI tools across various scientific domains.}




\maketitle

\section{Introduction}\label{sec1}
Achieving precise synthesis has long been a dream for materials chemists. This involves using a range of controllable material synthesis techniques to create materials with specific structures and properties based on the underlying physicochemical mechanisms. \cite{Li2014AnisotropicGN, Yang2022BigDI, Sun2002ShapeControlledSO, tang2020machine} To overcome the limited scope of each set of synthetic conditions, building connection across a wide range of methods and scenarios is crucial. \cite{szymanski2021toward, Wang2023ScientificDI} This would expand the feasibility and adaptability of synthetic processes, ultimately enabling the tailored production of materials to meet specific scientific and technological requirements. \cite{Xia2009ShapecontrolledSO, Wang2010SimultaneousPA, Xing2009HighlyCC}

Standing at the forefront of the times, designing cutting-edge deep learning methods combined with existing knowledge is one of the most promising methods to achieve controllable material synthesis. \cite{So2020DeepLE, Choudhary2021RecentAA, tao2021nanoparticle} It is important to note that all literature are written in human languages. In this context, large language models (LLMs), such as GPT-4, are promising solution to complex problems. It has demonstrated exceptional results in autonomous biological and chemical synthesis experiments, amongst other domains, because of their learning ability. \cite{Achiam2023GPT4TR, Xiao2023GenerativeAI, AI4Science2023TheIO, Zheng2023AGR, Guo2023WhatCL, Bran2023ChemCrowAL, pyzer2022accelerating} Despite efforts to let LLMs deal with synthesis tasks, a critical question remains: do these LLMs grasp the realworld physicochemical principles? Solid foundations or mere conjectures?

\begin{figure}
	\centering
	\includegraphics[width=1.0\textwidth]{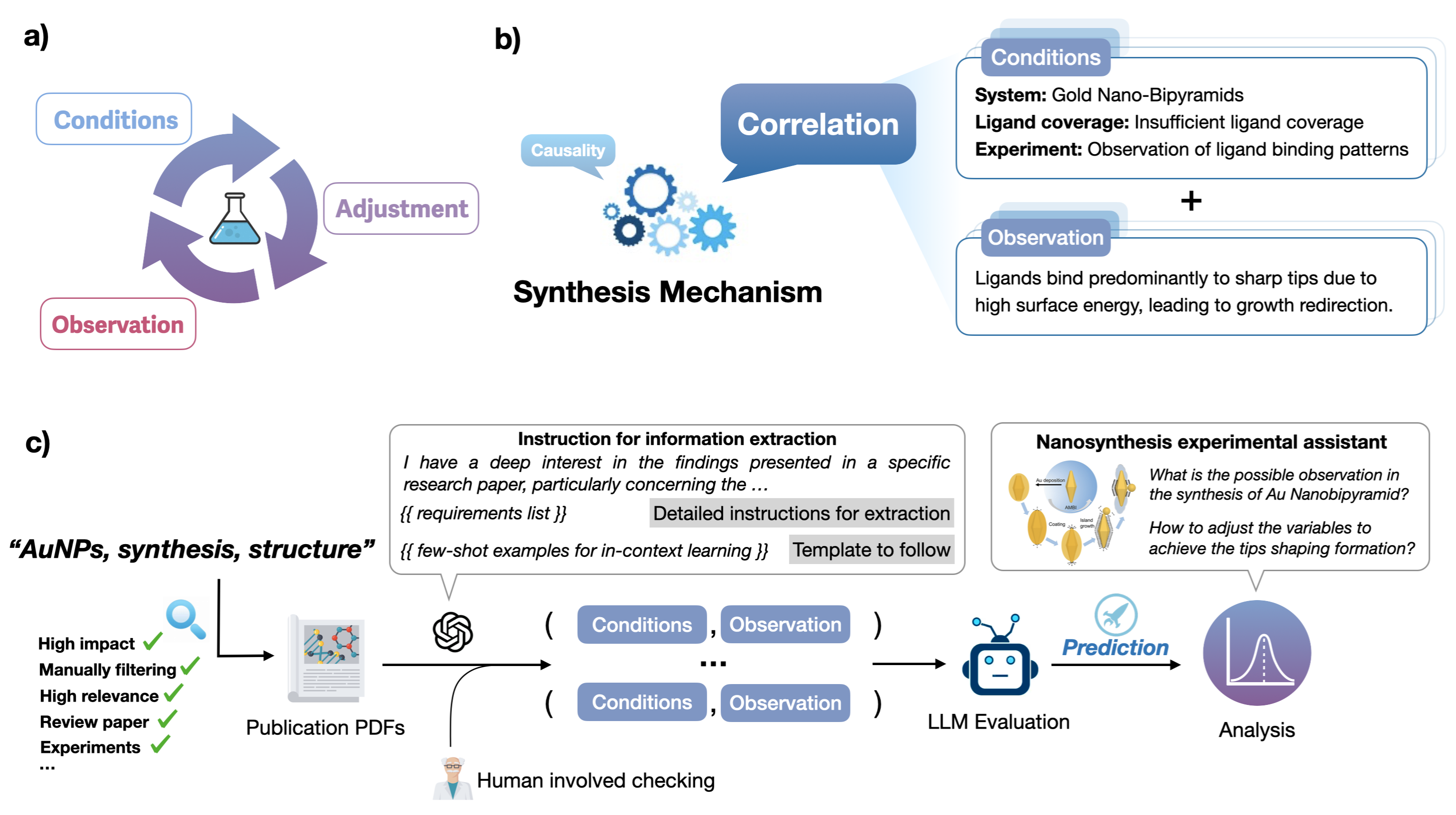}
	\caption{\textbf{Semantic illustration of our proposed framework for large language model evaluation in nanomaterial synthesis prediction, highlighting concepts and workflow.} a) Nanosynthesis study loop: begins with basic conditions, leading to the discovery of novel synthesis rules through experiments involving variable adjustments. b) exemplifies  the synthesis mechanism, dissected into causality and correlations, with an emphasis on correlations described through condition-observation pairs. c) outlines the process from sourcing relevant literature (using key area keywords) for benchmark construction and model evaluation.}
	\label{fig:framework}

\end{figure}

Among existing investigations, the common and straightforward approach for answering this question is fact-based evaluation, which can measure the learning performance of the model. \cite{weston2019named, cruse2022text, venugopal2021looking, zaki2024mascqa, jablonka202314, zheng2023chatgpt, kang2023chatmof, jablonka2024leveraging, guo2023can, zheng2023large, thawani2020photoswitch}. Meanwhile, evaluating the cognitive logic behind the principles is far more challenging yet essential for addressing key scientific issues.  \cite{chang2024survey} For instance, Alexander Fleming observed that bacteria could not survive where mold grew—a simple fact. Yet, this correlation alone could not explain why the mold inhibited bacterial growth. Through reasoning, Fleming discovered the penicillin. This case underlines the critical role of reasoning in scientific research.

Inspired by the research of nanomaterial synthesis, we embark on a feasibility study regarding whether LLMs can truly comprehend underlying physicochemical principles. Specifically, we construct 775 expert-level test questions, covering six primary methods of gold nanoparticle synthesis and six major categories of nanomaterial structures, to thoroughly assess the capabilities of current LLMs, more details are shown in Method section. In this work, our contribution encompasses the following key elements, also shown in Figure \ref{fig:framework}:

\begin{enumerate}
	\item We propose a synthetic mechanistic descriptor, grouped by initial conditions, variable adjustments, and experimental observations, to deal with the material synthesis mechanism study.
	\item A benchmark of gold nanoparticle synthesis mechanisms, including 775 multiple selection questions focusing on synthesis experiments, is built by using the descriptor.
	\item A confidence-based score (c-score) is introduced, offering an interpretable measurement of LLMs to understand complex synthesis mechanisms.
\end{enumerate}

\section{Method}
\subsection{Preparation of datasets}
\begin{figure}
	\centering
	\includegraphics[width=1.0\textwidth]{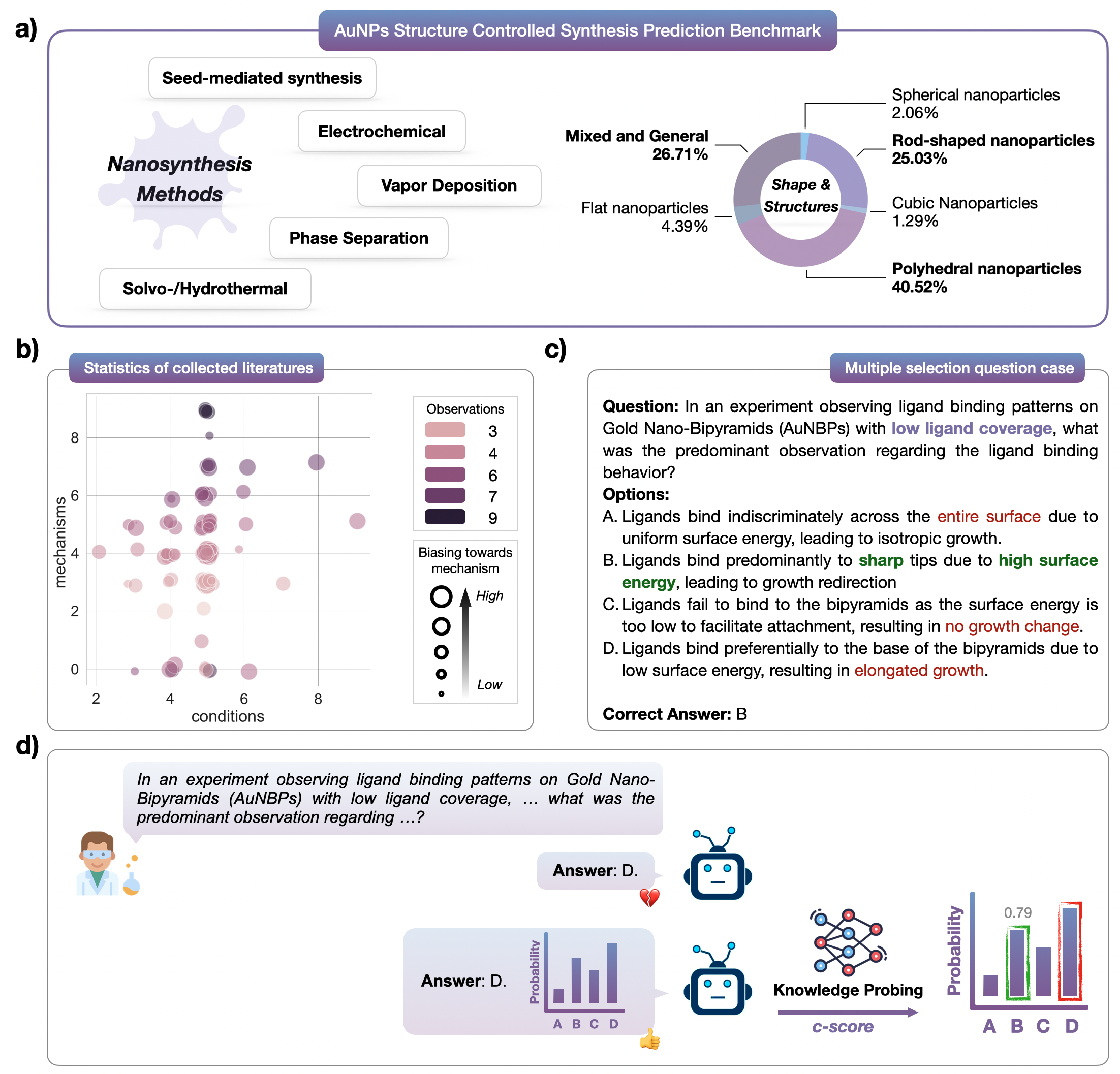}
	\caption{\textbf{Evaluation data set illustration.} a) shows the distribution of collected evaluation sets containing 775 questions categorized by synthesis methods and structures, respectively. b) displays a jittered scatter plot of manually curated research papers with the counts of mechanism, conditions and observations, with mechanism relevance from low to high, indicated by varying colors to represent the frequency of observations and varying sizes to represent the biasing towards mechanism. c) showcases the multiple selection question considered in the evaluation. The model is instructed to give the correct option. d) illustration of the probing test in our evaluation study based on the proposed c-score.}
	\label{fig:datasets}
\end{figure}

The dataset for evaluation was meticulously selected from high-quality papers boasting an IF $>$ 15, with a particular emphasis on synthesis methods for controlling the structure of gold nanoparticles. To ensure the relevance and quality of the papers to meet the theme of gold nanoparticle synthesis, we manually reviewed over 220 articles from a diverse range of publishers and esteemed scientific journals. Each paper underwent a process of key experimental information extraction, leading to the summarization and collection of 775 experimental records, categorized under a condition-observation-mechanism fashion. This structure serves as a descriptor for expressing any controlled experiment and is a usual loop for discovering synthesis principles. This systematic design not only enabled the creation of a unified framework of descriptors to assist in compiling test questions, but also, given the unstructured nature of mechanistic expression in nanomaterial synthesis, emphasized the importance of using average sampling in the design of the evaluation set. In this approach, the dataset distribution and cases are shown in Figure \ref{fig:datasets}, and we address three pivotal issues and challenges:

\begin{enumerate}
\item The keywords classification of evaluation content and the synthesis of knowledge for uniform sampling in this domain.
\item The precise extraction of conclusions and the completeness of experiment conditions in each report.
\item The completeness of correlation between experimental conclusions and mechanisms of nanomaterial synthesis.
\end{enumerate}

Regarding the keywords classification issue, our evaluation perspective is based on either material synthesis methods or material morphology. The former category predominantly focuses on the seed-mediated synthesis method, a commonly utilized approach for synthesizing nanoparticles with complex structures. The latter category emphasizes mechanisms more generally applicable to nanocyrstals, rod-shaped particles, and some unmarked systems, with their distribution illustrated in Figure \ref{fig:datasets}a. Two distributions of the data indicate a wide range knowledge points of gold nanoparticle synthesis considered in this work.

For the insight extraction problem, we employed cue word engineering in conjunction with predefined descriptors to extract the descriptions of synthesis experiments from each paper. This was achieved by using pre-defined prompts and then, manual checking. It was determined that for each report, on average, experimental report included five particularly relevant initial conditions and three sets of experimental setups, such as the increase or decrease of certain parameters, the presence or absence thereof, along with a corresponding number or more of observations leaning towards conclusions, as shown in Figure \ref{fig:datasets}b, each point represents a scientific report. To facilitate easier visualization, we applied a jittering technique, meaning that each point was moved slightly by adding a small random values to the integer coordinates.

For mechanism completeness, we posit that accurate division based on mechanistic tendencies can significantly enhance the precision of model evaluation. However, considering the inherent bias in dividing mechanisms—where a handful of studies fail to report complete and comprehensive descriptions of mechanisms compared to the average, and there exists a wide variance in the interpretation of complete mechanisms. To illustrate this point, we employ the GPT-4 via OpenAI API to model in conjunction with cue words for a more detailed division of mechanistic tendencies. Previous research has indicated that there is no large gap in evaluation between humans and LLMs even in the specific area of material science. Meanwhile, we also confirmed that 775 questions are almost making the accuracy converged for evaluating LLMs according to their performance in this task, see supplementary information (Figure S1) for more details. Furthermore, we discovered that the utilization of cue word engineering with the GPT-4 model enables rapid acquisition of summaries for nanomaterials papers, which is also demonstrated in existing work. \cite{clark2021all, chiang2023can, Zheng2023AGR}

Furthermore, we discovered that the utilization of cue word engineering with the GPT-4 model enables rapid acquisition of summaries for nanomaterials papers, which is also demonstrated in another work. \cite{rampal2024single} This capability extends even to articles that do not contain experimental data (such as literature reviews, opinions, and comments), wherein GPT-4 provides feedback indicating the absence of extractable content, thereby demonstrating GPT-4's honesty in responding to user queries.

Ultimately, we can rephrase this refined condition-observation pair-wise data into a standardized format of questions and options with the gold answer using the GPT-4 along with predefined instructions, one case is shown in Figure \ref{fig:datasets}c. These are formatted as multiple-choice questions with four options. Notably, sampling analyses have shown that the powerful paraphrasing capabilities of GPT-4 enable it to convert these data into equivalent test questions smoothly. Furthermore, we can filter, adjust and summarize them through modifications of cue words as deemed appropriate. For detailed methodologies on the aforementioned cue word engineering, and prompt with instructions, please refer to the supplementary information (Note S1).

\subsection{LLM Baselines}
We choose multiple existing models with different architectures and features for comparison, to better consider the inner design differences.

\textbf{Vicuna Paradigm.}  Vicuna represents a pioneering effort within the open-source community. This model, conceptualized and refined by LMSYS, undergoes an extensive fine-tuning process leveraging the LLaMA series models \cite{touvron2023llama}, trained on a dataset comprising 70,000 user-generated dialogs. Furthermore, Vicuna is distinguished as one of the preeminent models within the subset of LLaMA-2 fine-tuned models (for vicuna-v1.3 and v1.5 versions), attributed to its superior training quality and the voluminous corpus of data it utilizes. \cite{vicuna2023}

\textbf{Mistral and Mixtral architectures. }  Developed by Mistral.AI, these models represent two distinct approaches within the field of AI. The Mistral model integrates a Grouped-Query Attention mechanism to enhance inference speed and employs Sliding Window Attention to efficiently manage extended sequences with reduced computational demands. Conversely, the Mixtral model adopts an innovative architecture characterized by a high-quality Sparse Mixture of Experts. This decoder-only framework enables the feedforward block to select from eight unique parameter groups, with a router network at each layer determining the optimal combination of two groups, known as experts, to process each token and amalgamate their outputs in an additive fashion. \cite{Jiang2023Mistral7, Jiang2024MixtralOE}

\textbf{Qwen series.} These models are meticulously fine-tuned using a dataset curated to align with a diverse range of tasks, including conversation, tool utilization, agency, and safety protocols. A notable distinction of the Qwen models lies in their token representation capacity. The 7B model processes 2.4 trillion tokens, while the 14B model handles 3.0 trillion tokens. This positions Qwen at the pinnacle of token representation capabilities compared to other models in its category. \cite{Bai2023QwenTR}

\textbf{Gemma framework.} The Gemma model encapsulates a series of lightweight, cutting-edge open-source models that draw from the same foundational research and technological advancements underpinning the Gemini models. This lineage of models is celebrated for their formidable performance metrics, notably in comparison to the GPT-4 model. \cite{team2023gemini}

\textbf{Other framewroks.} In the evaluation of models including, but not limited to, GPT-4 (gpt-4-0125-preview) and Claude 3 (claude-3-ops-20240229), our analysis endeavors to assess them based upon their sophisticated capabilities in solving general problems. It is pertinent to note that these models are not made available as open-source; nevertheless, they are developed through the training on extensive corpuses of data, encompassing a wide array of domains. This approach underscores the depth and breadth of knowledge these models can potentially harness, despite the proprietary nature of their development methodologies. In addition, Gemini is not considered due to its accessibility, thus, we use Gemma for a case to test its behavior.  \cite{Achiam2023GPT4TR}

\subsection{Evaluation metrics}
The most direct metric for evaluation is the use of accuracy to assess the comprehensive performance of models across a certain number of test questions, with higher scores indicating stronger capabilities. This is widely used in various tasks, and benefits from a design similar to that of comprehensive human knowledge tests, directly aiming at the logical reasoning and knowledge understanding abilities of language models through a multiple-choice question format. \cite{Laskar2023ASS, singhal2023large}

In this work, we aim for the model to answer the question with confidence, reflecting a quantitative understanding of the intrinsic sequential logic in material synthesis. Thus, we further evaluate LLMs by introducing the c-score with knowledge probing techniques, and more designing details are shown in the results section. Here we treat statistical accuracy, with the counting of true-or-false, as a baseline metric to evaluate the comprehensive capabilities of models, to explore their performance relative to random guessing.

To sum up, two perspectives with the proposed benchmark are illustrated regarding whether LLMs understand physicochemical principles, i.e., true-or-false-based accuracy, and ensure the discernment of correct answers, i.e., confidence-based score, as shown in Figure \ref{fig:datasets}d. To achieve this, with a preliminary study of temperature effects on LLMs for examining the degree of stability, we use both intuitive accuracy and c-score to judge the capabilities of models in recognizing physicochemical principles.

\begin{figure}
	\centering
	\includegraphics[width=1.0\textwidth]{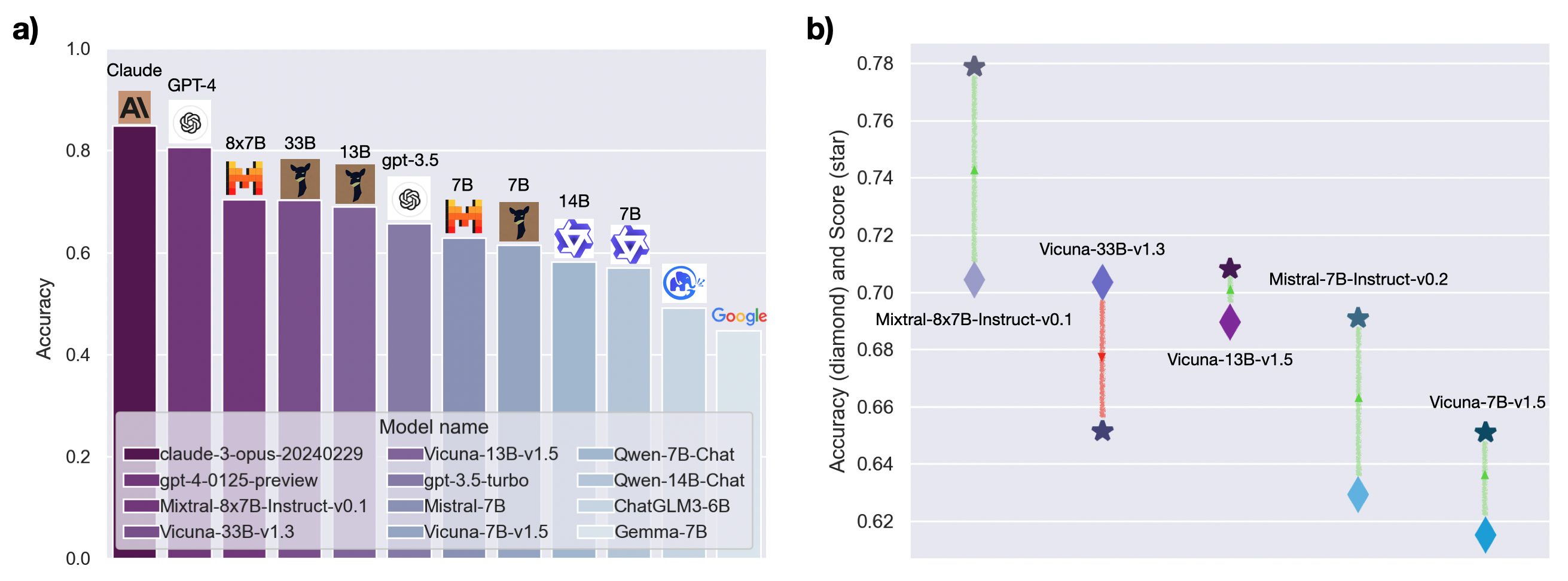}
	\caption{\textbf{Evaluation accuracy of baselines and the confidence-based scores of top-5 open-sourced models compared to the original accuracy in multiple selection questions. } a) x-axis represents different models, while y-axis is the accuracy. The figure delineates the range in accuracy achieved by each model under different temperature settings (from 0.0 to 1.0), where the circles represent the accuracy at each temperature setting, and the diamonds denote their average values.  b) The comparison between accuracy and condifence-based scores among 5 top-performance models, showing the performence increasing (green line) and decreasing (red line). }
	\label{fig:results_acc_and_score}
\end{figure}

\section{Results}

\subsection{Temperature Effect Analysis}
\begin{figure}
	\centering
	\includegraphics[width=1\textwidth]{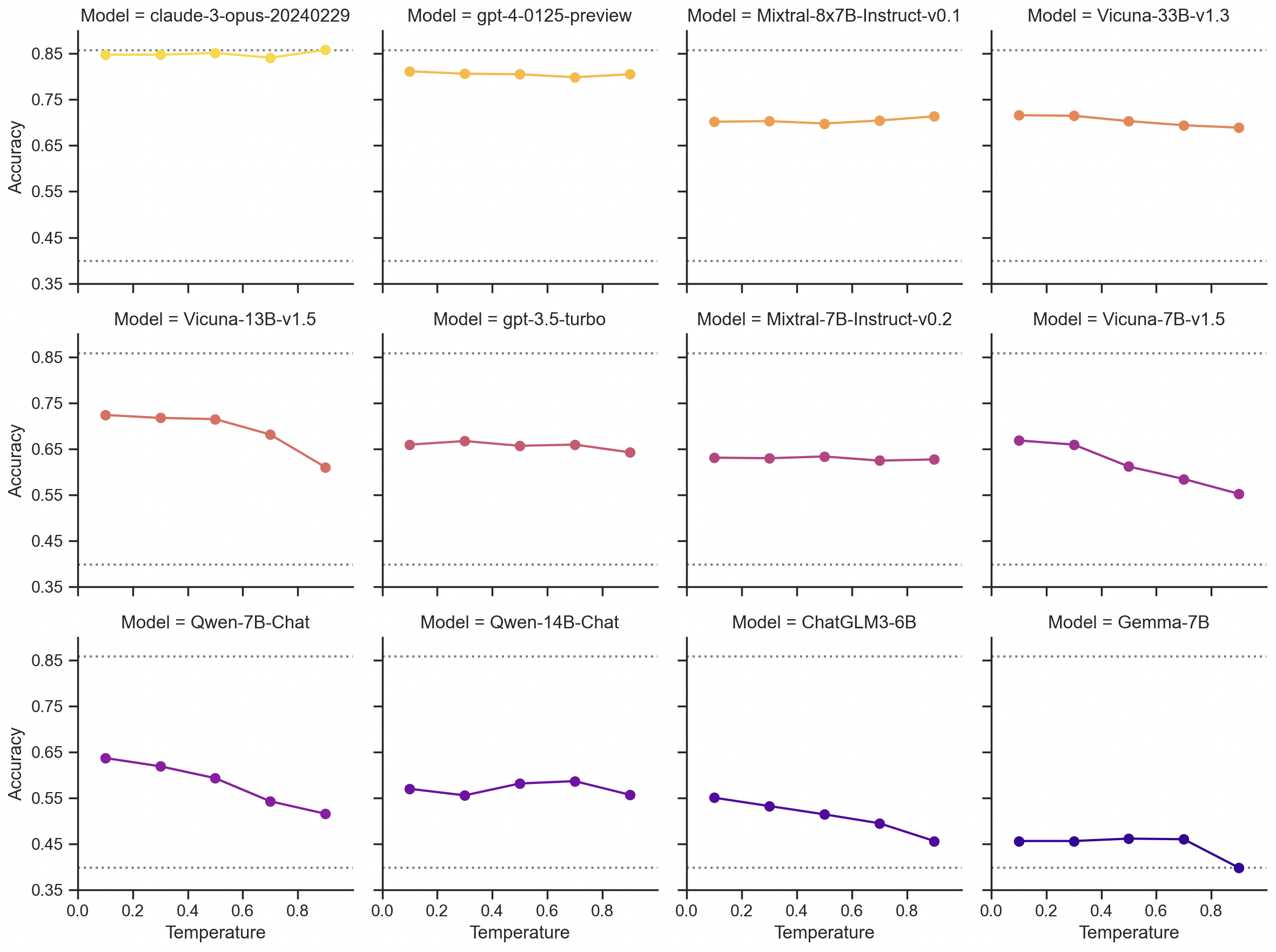}
	\caption{\textbf{Evaluation results of temperature effects on baselines}. Each subplot illustrates the accuracy (y-axis) of the corresponding model under various temperature settings (x-axis), organized in descending order based on the average of the model performance across multiple-choice questions at different temperatures. Each point denotes the accuracy at a fixed temperature.}
	\label{fig:result_average}
\end{figure}

Language models based on the transformer architecture predict the next token through an autoregressive approach and iteratively generate the output for an entire sentence. This method also allows for the adjustment of the probability distribution of the final predicted token.

The decision to adjust the probability distribution using the temperature setting in language models is inspired by statistical thermodynamics, where a higher temperature signifies a greater likelihood of overcoming energy barriers. In probability models, logits function as a representation of kinetic energy. Low temperatures lead to a more concentrated distribution of values, whereas higher temperatures yield a dispersed distribution. The introduction of temperature into logits facilitates temperature sampling, which, upon being fed into the Softmax function, yields sampling probabilities.

To examine the degree of stability, we investigated the impact of different temperatures on the performance of language models. We uniformly selected five temperature values ranging from 0 to 1, specifically 0.1, 0.3, 0.5, 0.7 and 0.9, to conduct controlled experiments. As demonstrated in the Figure \ref{fig:result_average}, there is a trend of precision decline in models as temperature increases, with certain models exhibiting more complex fluctuations. For instance, both claude-3-ops-20240229, gpt-4-0125-preview, Mixtral-8x7B-Instruct-v0.1, Mistral-7B-Instuct-v0.2 and gpt-3.5-turbo showed a trend where accuracy initially decreased and then increased as the temperature slightly rose.

From the final outcomes, it is evident that temperature exerts a patterned influence on the performance of language models. Although there remains a possibility of incorrect responses, extensive statistical evidence suggests that their overall performance is marginally superior to that observed at higher temperatures, demonstrating an assured stability.

\begin{figure}
	\centering
	\includegraphics[width=0.65\textwidth]{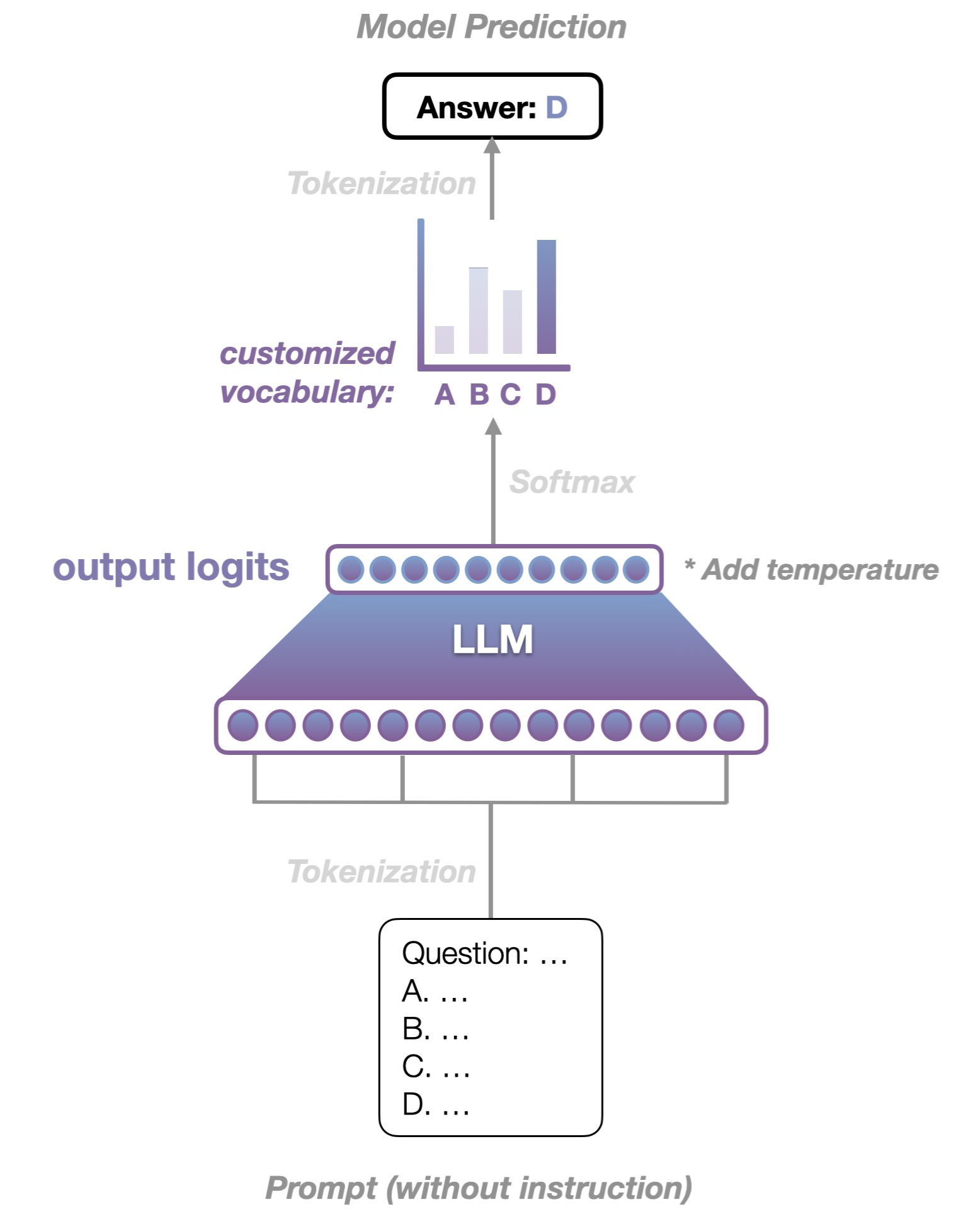}
	\caption{\textbf{Illustration of the knowledge probing method.} Given the input with both question-options and instructions, the model should give answer with predefined vocabulary, which includes A, B, C and D. The probability of each option is drawn based one the  ouput logits with fixed setting of the temperature before Softmax. By observing the distribution changes of all options, model behaviors can be revealed upon different test cases. We use this design to test models ability regarding the knowledge of AuNPs synthesis. }
	\label{fig:intro_probing}
\end{figure}

\subsection{Results of Accuracy with Temperature}
It is widely recognized that the capabilities of language models are predicated upon two distinct phases: pre-training and fine-tuning. On one hand, the method of pre-training is acknowledged to endow models with an understanding of context. Since it necessitates that language models predict masked words by comprehending the context (or the surrounding words), to elucidate, when describing a dog, one needs to employ descriptions such as a tailed mammal, mankind's best friend, etc., leveraging context to grasp the meaning of dog. On the other hand, the fine-tuning process equips language models with a degree of obedience to instructions and the ability for sustained dialogue, with its efficacy contingent upon the volume, variety of the final dataset, and the process of the model fine-tuning.

Contemporary large language models rely on the aforementioned pre-training and fine-tuning learning processes. Consequently, multiple-choice questions serve as one of the effective methods to evaluate the level of reasoning ability of the language model in a specific domain. This implies that the language model needs to provide answers according to the domain knowledge learned during the pre-training phase and the comprehensive abilities acquired during the fine-tuning phase, as required by the question.

The content we aim to evaluate primarily unfolds from two aspects:

\begin{enumerate}
	\item The themes encompassed by the multiple-choice questions represent the knowledge being assessed, aiming to evaluate the understanding of the model in terms of the concepts and semantics demonstrated within the sentences.
	\item This question format supports both the different mechanisms expression and the examination of the language model's logical reasoning abilities, assessing whether it can answer based on the fundamental principles of gold nanoparticle synthesis.
\end{enumerate}

In order to obtain the binary accuracy of the model, for each multiple-choice questions, one point will be given to the model if it selected the gold answer, otherwise zero. This process will also be repeated with different temperature settings, i.e., from 0.1 to 0.9 with 5 steps. Finally, the average score of each model will be ranked.

As illustrated in Figure \ref{fig:results_acc_and_score}a, all models significantly surpass the random guessing baseline of 25\% with a remarkable margin, consistent with their capabilities in general tasks.

It is worth noting that Claude and GPT-4 are the best performing models in this evaluation, with accuracies of 84.8\% and 80.5\%, respectively. The other open source models are not very competitive, with accuracies around 70\% or lower, showing a huge gap with the top two. Specifically, Claude and GPT-4 have outstanding performance on a wide range of benchmarks due to their excellent training. The open source models may perform relatively poorly due to a variety of reasons such as the type and quality of training data and model size. Among them, the Mixtral-8x7B model performs slightly better than all other open-sourced models tested, with an accuracy of about 70.4\%. Gemma has an accuracy of about 44.7\%, which is the last one in our evaluation. However, its accuracy is still much higher than 25\% (random guess). In addition, the accuracy of other models is in the middle level, their accuracy values can be found in the supplementary information (Table S1). To sum up, we believe that all of these tested LLMs can explain some physicochemical principles, whether basic or complex, showing future potential in explaining synthetic mechanisms.

\subsection{Results of C-scores with Temperature}

\begin{figure}
	\centering
	\includegraphics[width=0.99\textwidth]{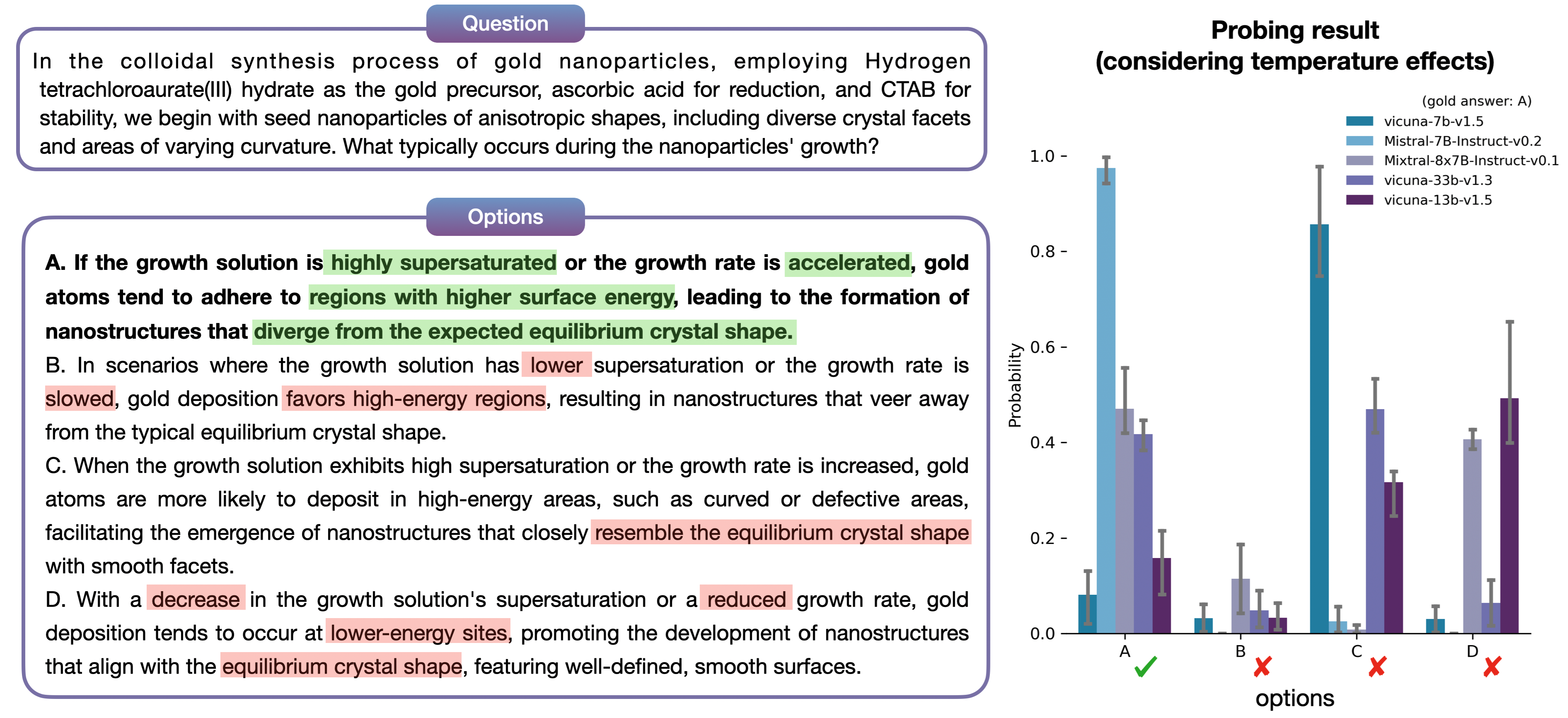}
	\caption{\textbf{Knowledge probing case.} The text on the left side displays the question and answer pair. The results indicate that LLMs assign different probabilities to the given options, with temperature effects represented as error bars.}
	\label{fig:probing_case}
\end{figure}

Knowledge probing is a method designed to assess the capacity of language models, such as those in the GPT series, to understand and recall specific knowledge domains. This technique evaluates the model's comprehension by analyzing the probability distribution of tokens corresponding to the logits in the model's output. Since the prediction of the next token is governed by the distribution of logits and transformed by the Softmax function, each token is assigned a probability. Our focus is directed towards a subset of tokens with higher probabilities rather than a singular output result, as illustrated in Figure \ref{fig:intro_probing}. Typically, a sharper distribution of predicted token probabilities indicates a higher certainty in the model response, and vice versa. This design enables an analysis of the model's responses to varied queries, discerning whether they are grounded in solid theoretical understanding or merely speculative guesses.

In our study, we examine the distribution at the logits layer before the model responses, which is considered an indicator of confidence, to better address the aforementioned question. We further assess this by combining previous accuracy metrics with c-score. In past multiple-choice assessments, the final answer of the model is assumed to be chosen with 100\% confidence, meaning that either True or False only. This inspires us to measure its capabilities directly by the percentage of the gold answer’s confidence, even the model chooses the drinkable gold answer. We here evaluate the overall confidence using the formulated c-score, which quantifies the confidence level assigned to each correct answer, as detailed in the E.q. \ref{eq:c_score}:

\begin{equation}
	\textit{\text{c-score}} = \frac{1}{N}\sum_{i=1}^{N}\frac{e^{L_G^i}}{e^{L_A^i} + e^{L_B^i} + e^{L_C^i} + e^{L_D^i}}
	\label{eq:c_score}
\end{equation}
where $L_G^i$ is the probability (or confidence) of gold answer regarding the i-th question and $L_X$ is for other options. The probability of all options (assume four here) is normalized exponentially and N is the number of all questions.

Specifically, based on the evaluation of accuracy, we further evaluate the top-5 ranked open-source models with c-score for efficient comparison, considering only models that excel in the benchmark. These models represent typical scales in the open-source community. The evaluation process will also be repeated with different temperature settings, as mentioned before (0.1, 0.3, 0.5, 0.7 and 0.9). Finally, the average c-scores of top-5 models will be ranked for discussion due to their competitive performances.

Regarding the results, as shown in Figure \ref{fig:results_acc_and_score}, the Vicuna-33B  exhibits a lower level of performance, whereas the other models showed slight improvements in the c-score compared to accuracy. In detail, the c-score of Mixtral-8x7B improves by about 8\% compared to the accuracy, while the c-score of Vicuna-33B decreases by about 6\%. This indicates a clear confidence difference between the two models. Similarly, Vicuna-13B, Mistral-7B and Vicuna-7B demonstrate remarkable differences between accuracy and c-scores. The results indicate that the c-score effectively measures the ability of LLMs in gold nanoparticle synthesis tasks in an interpretable manner. Such metrics suggest that utilizing c-score allows for a more appropriate assessment of language models compared with pure accuracy, revealing insights distinct from traditional accuracy statistics.

For a direct view of the knowledge probing, we showcase a probing result in Figure \ref{fig:probing_case}, where we consider the temperature effects, which are represented by the error bar. Here, option A is correct, because the higher the supersaturation (reduction rate), the more unbalanced the growth of the gold particle morphology will be, and thus a nanoparticle morphology with a high-index crystal face or high curvature will be developed. Options B, C, and D have opposite statements. For each model tested, except for option B, there is a tendency to choose A, C, and D -- with a higher confidence. Among them, Mistral-7B has a confidence of nearly 100\% for the correct option A, and a tendency for other options is almost 0. This result shows that the model has sufficient and solid learning of this knowledge point, and can distinguish the nanosynthesis logic involved in this question, while other models are more confused. One possible reason is that the model is interfered by certain keywords, resulting in a confidence level of about 50\%. Some other examples of knowledge probing are in supplementary information (Note S2).

\section{Related Work}

In the field of materials science, existing datasets predominantly support tasks focused on factual knowledge, such as named entity recognition and classification. \cite{weston2019named, cruse2022text, venugopal2021looking, kim2017materials}	Researchers utilize these datasets to benchmark the performance of language models in the materials domain.

Previously, three key chemistry-related capabilities in LLMs, understanding, reasoning, and explaining have been identified, and a benchmark containing eight chemistry tasks has been established.  \cite{guo2023can}  Meanwhile, the potential of large language models to perform scientific synthesis, inference, and explanation across many domains for scientific discovery has been discussed, although this approach is based solely on knowledge graph inference. \cite{zheng2023large} To expend the task diversity, LLMs such as GPT-3 have been benchmarked on datasets spanning the chemical space, including molecules, materials, and reactions, across diverse tasks such as classification, regression, and inverse design. \cite{jablonka2024leveraging} With the continuous growing of the LLMs, a dataset of 650 challenging questions from the materials domain, requiring the knowledge and skills of a materials science student who has completed their undergraduate degree, has been curated. \cite{zaki2024mascqa}

While significant progress has been made in benchmarking, there remains a need for more comprehensive evaluations that encompass the full spectrum of capabilities required for advanced scientific applications. This includes the ability to reason about mechanisms and the fundamental rules of physics and chemistry. Our study developed a benchmark consisting of 775 multiple-choice questions focusing on the mechanisms of gold nanoparticle synthesis and propose a novel evaluation metric, the confidence-based score (c-score), which probes the output logits to derive the precise probability for the correct answer.

\section{Conclusion}
In this study, we introduced a novel evaluation method for assessing LLMs in the context of materials science, specifically focusing on the synthesis of gold nanoparticles. Our approach encompassed the development of 775 multiple-choice questions addressing both synthesis methods and morphological structures. By employing knowledge probing and confidence scores (c-scores), we evaluated a range of mainstream open-source and closed-source LLMs. The results of our evaluation demonstrate that c-scores are more effective in discerning whether LLMs' contributions to the synthesis tasks are rooted in an understanding of the physicochemical mechanisms, rather than mere recall of information. This finding underscores the importance of assessing the models' comprehension and logical reasoning abilities, which are crucial for facilitating genuine scientific discoveries. In conclusion, our study not only highlights the potential of LLMs in advancing materials science but also sets a precedent for the rigorous evaluation of their scientific and logical reasoning capabilities. The insights gained from this research can inform the development of more sophisticated models that are capable of making meaningful contributions to scientific discovery.

\section*{Data and code availability}

The dataset created for this study, along with the testing code, is available in the GitHub repository at \url{https://github.com/Dandelionym/llm_for_mechanisms.git}. This repository contains all relevant data and scripts necessary to replicate our experiments and results.


%
%
%
%
%
%
%
%
%
%
%
%

\end{document}